\documentclass[a4paper,12pt]{article}
\usepackage{array}
\usepackage{geometry}
\usepackage{caption}
\usepackage[utf8]{inputenc}
\usepackage{amsmath}
\usepackage{graphicx}
\usepackage{hyperref}
\usepackage{float} % H seçeneği için float paketini ekleyin
\title{LLMs-in-the-loop Part-1:
Expert Small AI Models for Bio-Medical Text Translation}
\author{Bunyamin Keles, Murat Gunay\footnote{Corresponding Author: joseph@aimped.ai}, Serdar I.Caglar \\ 
\\AI Amplified (Aimped) Inc.,\\
1775 Tyson Blvd, Floor 5 Tysons, VA 22102 \\ 
\\\texttt{\{benjamin, joseph, forest\}}@aimped.ai}

\begin{document}

\maketitle

\begin{abstract}
\noindent Machine translation is indispensable in healthcare for enabling the global dissemination of medical knowledge across languages. However, complex medical terminology poses unique challenges to achieving adequate translation quality and accuracy. This study introduces a novel "LLMs-in-the-loop" approach to develop supervised neural machine translation models optimized specifically for medical texts. While large language models (LLMs) have demonstrated powerful capabilities, this research shows that small, specialized models trained on high-quality in-domain (mostly synthetic) data can outperform even vastly larger LLMs.\\

\noindent Custom parallel corpora in six languages were compiled from scientific articles, synthetically generated clinical documents, and medical texts. Our LLMs-in-the-loop methodology employs synthetic data generation, rigorous evaluation, and agent orchestration to enhance performance. We developed small medical translation models using the MarianMT base model. We introduce a new medical translation test dataset to standardize evaluation in this domain. Assessed using BLEU, METEOR, ROUGE, and BERT scores on this test set, our MarianMT-based models outperform Google Translate, DeepL, and GPT-4-Turbo.\\

\noindent Results demonstrate that our LLMs-in-the-loop approach, combined with fine-tuning high-quality, domainspecific data, enables specialized models to outperform general-purpose and some larger systems. This research, part of a broader series on expert small models, paves the way for future healthcare-related AI
developments, including deidentification and bio-medical entity extraction models. Our study underscores the
potential of tailored neural translation models and the LLMs-in-the-loop methodology to advance the field
through improved data generation, evaluation, agent, and modeling techniques.\\

\end{abstract}

\section{Introduction}
The ability to think and communicate is significantly impacted by language, which has become increasingly
important due to technological advancements and global internet access. English has emerged as a widely used
lingua franca, with numerous individuals worldwide learning it to facilitate communication \cite{al-sobhi2018,jenkins2017}. In societies
where multiple languages are spoken, effective communication is crucial for cross-border interactions and
daily exchanges \cite{anderson2018}. As a result, there is a growing demand for translation, which plays a vital role in academic
language teaching and enables the global dissemination of information \cite{anderson2018,kim2013}. However, translating poses
challenges as it involves transferring meaning between different languages, and neglecting the diverse range
of text types can hinder the accuracy and variety of translations \cite{al-otaibi2022}.\\

Medical translation is crucial in bridging communication gaps in the healthcare field. Throughout history,
languages such as Latin, Greek, Arabic, Hebrew, Syriac, and Persian have been translated for medical
education \cite{fischbach1998,mcmorrow1998}. English emerged as the dominant language in medical education during the 19th century, and
today, it is recognized as the universal language for scientific communication, with most medical research
articles written in English. This is especially important considering the vast number of medical articles
published annually, which amounts to approximately 10 million \cite{kaplan2001,montgomery2009,salager-meyer2014}.\\

Medical translation plays a significant role in knowledge transfer across various language domains, encompassing drug prospectuses, medical books, patient notes, and articles \cite{al-jarf2018,alasbahy2023,haddad1997}.With the widespread use of English in medical education and the preference among doctors in multilingual Arab countries, accurate and efficient translation of medical texts has become increasingly essential for interprofessional communication, prescription writing, and report generation \cite{alasbahy2023,medlinkstudents2023}. Particularly amid the global concern of the coronavirus
pandemic, the demand for comprehensible medical information in English has grown exponentially, highlighting the importance of human translators in improving access to healthcare and enhancing telemedicine practices \cite{dusek2014,yu2021}.\\

Due to the high cost and time constraints associated with expert translation, there is a growing need to develop affordable, high-quality, and accessible machine translation (MT) solutions in the medical field \cite{karliner2007,randhawa2013}. Effective medical translation is crucial in knowledge sharing and can significantly improve healthcare outcomes \cite{wolk2015}. Although English is widely used in medicine, many medical studies are written in languages other than English, necessitating medical text translation between English and other languages.\\ 

In the medical field, professionals use various types of texts to communicate with each other, such as discharge summaries, medications, case studies, case notes, epicrises, academic studies, and imaging reports. These texts contain specific terms and require accurate and prompt translation to avoid errors.\\

The advancement of artificial intelligence algorithms has paved the way for developing machine translation (MT) systems that can effectively address medical translation requirements \cite{koehn2007}.\\

These systems leverage the analysis of paired text samples and can be trained to cater to specific medical domains, enabling the generation of accurate translations \cite{koehn2007}. This study focuses on developing medical text translation models using the Neural Machine Translation (NMT) approach, which has gained attention for its ability to optimize translation performance \cite{bahdanau2015}. NMT employs a sequence-to-sequence (seq2seq) architecture with Recurrent Neural Network (RNN) components, consisting of an encoder and a decoder, to convert source text into target text in different languages \cite{cho2014,sutskever2014}. To handle longer sequences, variants of RNN cells such as Long Short-Term Memory (LSTM) or Gated Recurrent Units (GRU) are utilized, ensuring relevant information transfer and enhancing translation accuracy over longer-term dependencies \cite{chung2014,hochreiter1997}.

\section{Background}
With the impact of globalization, interaction has increased in every field, such as medicine. In this context, accurate and effective translation of medical texts between different languages is vital for disseminating research findings, sharing clinical data, and overcoming language barriers so scientists can work in parallel with their colleagues in other nations.\\

Users may use a translation to get a general idea of a context. However, translation accuracy is crucial in some applications, such as medicine, where zero errors are required. A mistranslation between patient and physician can jeopardize patient safety. While Statistical Machine Translation (SMT) progress is slowing down, promising methods such as neural networks are needed \cite{costa-jussa2012}. Machine translation has also been applied to the medical field due to advances in language technologies. For example, in the US, both regional and national translations of medical documents have been carried out \cite{kirchhoff2011}. While these translations were previously considered insufficient, developing technologies show they can soon provide translations of the required quality \cite{wolk2015}.\\

Machine Translation (MT) is closely related to advanced rule-based systems or statistical phrase-based methods and dates back many years \cite{stein2018}. NMT, first introduced in 2013 \cite{kalchbrenner2013}, has become more popular recently \cite{pilch2022}. In this context, a new neural network model called RNN Encoder-Decoder was proposed by researchers in 2014 \cite{cho2014}. In 2016, Google developed the Google Neural Machine Translation (GNMT) model using deep learning techniques \cite{coldewey2017,wu2016}, which the German company DeepL developed as a CNN-based model in 2017. Today, open-source translation solutions trained by Hugging Face have paved the way for researchers to translate independently \cite{pilch2022}.\\

In this study, we present the results of medical translation models by fine-tuning MarianMT, which has recently been very popular in MT models, using medical texts from six different languages.\\

MarianMT is an efficient and self-contained NMT framework written entirely in C++ with minimal dependencies. It relies only on Boost, CUDA, or a BLAS library, allowing barrier-free optimization at all levels. MarianMT features its automatic differentiation engine based on dynamic computation graphs. This enables implementing new models, custom operators, and GPU kernels without changing the core library. Optimizations like meta-algorithms for multi-node training, efficient batched beam search, and compact model implementations can be done in performant C++ code. MarianMT demonstrates state-of-the-art results on the WMT2017 English-German news translation task. It also shows case studies for automatic post-editing and grammatical error correction research. Experiments exhibit MarianMT’s high training and translation speeds, achieving 30x faster training than Nematus on 8 GPUs. MarianMT is a high-performance, flexible, selfcontained NMT framework allowing C++-level optimization. Its efficiency makes it suitable for cutting-edge NMT research \cite{junczys-dowmunt2018}.\\

Recently developed Large Language Models (LLM) have been used in many fields. Moslem, Haque \cite{moslem2023} showed that large language models such as GPT-3.5 can be used for adaptive MT. Their study indicates that GPT-3.5 can produce more consistent translations by imitating these examples when given sample translation sentences. It is also shown that GPT-3.5 can successfully perform tasks such as terminology extraction and terminology-constrained translation with zero training. The results show that GPT-3.5 can produce quality translations that compete with robust decoder-encoder-based MT systems. The work of Moslem, Haque \cite{moslem2023} is essential because it shows that large language models have great potential in adaptive MT. There are services on the web that provide acceptable quality translations like Moslem, Haque \cite{moslem2023} did. However, these translation services perform lower than their general performance in domain-specific translations such as medical areas.\\

It is known that in some cases larger models with more parameters, such as LLMs, do not produce better scores. In this context, in a study \cite{han2023}, it is stated that the Marian model \cite{junczys-dowmunt2018} with 7.6 mn parameters produces better scores using the ClinSpEn-2022\footnote[1]{https://temu.bsc.es/clinspen/} benchmark dataset compared to much larger models such as ClinicalNLLB \cite{costa-jussa2022} (54 bn) and Clinical-WMT21fb (4.7 bn) \cite{tran2021}.\\

Within the scope of this study, results are obtained with the AI Amplified MT model using the benchmark dataset used by Han, Gladkoff [35] and presented in the result section of this study. It is determined that the results we obtained are above the scores of all three studies given by Han, Gladkoff \cite{han2023}.

\section{Methodology} 
Medical texts are generally characterized by the unique features of medical terminology and meaningful and long sentences in different fields. The primary objective of this study is to achieve zero-error translation of medical texts with grammatical features such as passive constructions and third person.\\

The literature survey shows that translation models have been developed by fine-tuning MarianMT in studies such as English-to-Malayalam \cite{randhawa2013,saji2022}, English-Ukrainian \cite{maksymenko2023} and improving the Norwegian Translation model \cite{hellebust2023}. However, it can be stated that there are few medical translation models developed in different languages, and there is still a great need for medical text translation models.\\

Within the scope of this study, medical translation models have been developed in six languages (Table \ref{tab:Table1}).

\begin{table}[H] %farklı ne kullanılabilir        
    \caption{Medical Translation Language Pairs}
    \centering   
    \renewcommand{\arraystretch}{1.2} % Satır yüksekliğini artır
    \setlength{\arrayrulewidth}{1.5pt} % Çizgi kalınlığını ayarla
    \begin{tabular}{c}        
        %\textbf{Table 1: Medical Translation Language Pairs} \\
        \hline
        \textbf{Languages “from” and “to” English} \\
        \hline
        Deutsch \\
        Turkish \\
        French \\
        Romanian \\
        Spanish \\
        Portuguese \\
        \hline
    \end{tabular}    
    \label{tab:Table1}
\end{table}

For evaluation, the "Blue Score," "Rouge Score," "Meteor score," and "Bert score" were used to compare the results of the models. In many MT models, these scores compare the results obtained.\\

The Bilingual Under Evaluation (BLEU) score is a metric used to evaluate a generated sentence against a reference sentence, where a perfect match produces a score of one and an ideal mismatch produces a score of zero \cite{papineni2002}. The Metric for Evaluation of Translation with Explicit ORdering (METEOR) score is an automated metric for MT evaluation based on a generalized concept of unigram matching between machine- and humangenerated reference translations \cite{banerjee2005}.\\

Recall-Oriented Understudy for Gisting Evaluation (ROUGE) is a recall-oriented metric that compares summary quality by counting the number of overlapping words or n-grams between the summary and the ideal outline \cite{ganesan2018,lin2004}. BERT score is an automatic evaluation metric for text generation. Analogously to common metrics, it computes a similarity score for each token in the candidate sentence with each token in the reference \cite{zhang2019}.

\subsection{Datasets}
This study aims to translate medical texts into six different languages, as mentioned before. Custom datasets synthetically generated, and publicly available datasets were also used in the model development. Domain experts evaluate and curate high-quality language pairs derived from in-house labeled data and synthetically generated content. The sentence pairs and token numbers of the developed models are shown in Table \ref{tab:Table2}. \\

Within the scope of this study, in addition to the data on the Opus.nlp \cite{opus2023} web page, scientific articles, and medical texts belonging to each language were scraped, and parallel corpora were prepared by us to be translated from English to the specified target languages and from these target languages to English. The scrapping process was based on the condition that the source language must have an English translation. Publicly available datasets published by language pairs were also used \cite{opus-dataset2023}.\\

\begin{table}[H]   
    \caption{Medical Translation Base Model, Sentence Pairs, and Token Numbers}
    \centering
%    \textbf{Table 2: Medical Translation Base Model, Sentence Pairs, and Token Numbers} \\
    \renewcommand{\arraystretch}{1.2} % Satır yüksekliğini artır
    \setlength{\arrayrulewidth}{1pt} % Üst ve alt çizgi kalınlığını ayarla
    \setlength{\tabcolsep}{5 pt} % Sütun genişliklerini daralt
    \fontsize{9}{11}\selectfont % Tablo yazı boyutunu 9 yapar
    %\tiny % Metin punto boyutunu 9 yapar
    \begin{tabular}{l l ccc ccc}
        \hline        
        \textbf{Models} & \textbf{Base Model} & \multicolumn{3}{c}{\textbf{Sentence Pairs}} & \multicolumn{3}{c}{\textbf{Token Numbers}} \\
        \cline{3-8}
        & & \textbf{Train} & \textbf{Val.} & \textbf{Test} & \textbf{Train} & \textbf{Val.} & \textbf{Test} \\
        \hline
        en-es-en & mt-en-es/mt-es-en & 85,439 & 2,004 & 2,005 & 124,533,333 & 294,835 & 319,815 \\
        en-de-en & mt-en-de/mt-de-en & 1,450,808 & 2,913 & 2,914 & 232,339,417 & 469,100 & 468,107 \\
        en-fr-en & mt-en-fr/mt-fr-en & 1,905,850 & 2,867 & 2,868 & 25,521,977 & 377,787 & 381,422 \\
        en-tr-en & tatoeba-en-tr/mt-tr-en & 852,402 & 1,712 & 1,712 & 55,426,674 & 109,203 & 112,049 \\
        en-ro-en & mt-tc-big-en-pt/mt-roa-en & 1,438,810 & 3,615 & 3,616 & 57,599,188 & 29,852,963 & 29,852,093 \\
        en-pt-en & mt-tc-big-en-pt/mt-tc-big-pt-en & 1,920,276 & 2,889 & 2,890 & 158,078,074 & 82,144,779 & 82,144,380 \\
        \hline
    \end{tabular}
    \label{tab:Table2}
\end{table}

OPUS is a continuously updated collection of translated texts from the web. In the OPUS project, data is made available to users in various formats to help transform and align free online datasets, add linguistic annotation, and make it easier to work with the data. The main goal of Opus is to support data-driven NLP, mainly statistical MT. For this reason, it provides parallel datasets in various formats that can be used to train standard MT models. However, OPUS provides the data "as is" without guarantees or warranties. All preprocessing and alignment in OPUS datasets are done automatically, and no manual corrections are made \cite{opus2023}. In addition to our in-house datasets, SciElo \cite{scielo2023}, Mespen \cite{mespen2023}, EMEA \cite{emea2023}, and ELRC \cite{elrc2023}datasets were also used in the language translation model studied in the context of the developed medical translation models.\\

\subsection{Experimental Setup and Metrics}
We evaluated the performance of our developed medical translation models using proprietary test datasets from AI Amplified. The evaluation metrics employed in this study include BLEU Score, BERT Score, METEOR Score, and ROUGE Score. We compared our results against those obtained from Google Translate, DeepL, and OPUS translation models across all language pairs.\\

For the English-to-German medical translation model specifically, we extended our comparison to include Claude-3. The Claude-3-Opus model achieved scores of 0.511 for BLEU, 0.745 for ROUGE, and 0.741 for METEOR on our test set. Due to practical constraints such as time and cost, this broader comparison including large language models was limited to this language pair (Table \ref{tab:Table2}).\\

To further assess the quality of our translations, we innovatively employed ChatGPT and Claude AI as impartial judges. These LLM-based judges compared the outputs of Google Translate against our developed MT models (Tables \ref{tab:Table7}, \ref{tab:Table8}). This approach was chosen over human translation to mitigate time and cost constraints while still providing valuable insights into translation quality.\\

For this comparative evaluation, we randomly selected 100 sentences from our test set. The prompts used to instruct the LLM judges in this MT evaluation process are detailed below.\\

\subsubsection{Claude-3-Opus prompt}

\texttt{<<} Human: You are a professional translator and interpreter specializing in healthcare and biomedical. You will compare and evaluate the two translations.\\

I want you to help me in assessing translations from English to Turkish. The translated texts are from the healthcare and biomedical domain. Your evaluation should focus on adherence to medical terminology, simplicity, clearness, and accuracy, and you are expected to assign a score ranging from 0 to 100 to each translation. You MUST think, evaluate yourself, and do not generate text. Your ultimate output MUST only consist of the evaluation scores for each translation only in a JSON format as in \texttt{<}JSON\texttt{>}\texttt{<}/JSON\texttt{>} XML tags.\\

The output format:\\
\texttt{<}JSON\texttt{>} \{"Model-1": score, "Model-2": score\} \texttt{<}/JSON\texttt{>}\\
Do you understand your task?\\

Assistant: Certainly, I understood. You will provide two different translations from English to Turkish, and I will evaluate as a medical translator expert and assign scores to each one by adherence to medical terminology, simplicity, consistency, and accuracy. The format will be a JSON file containing only scores. Ok, let us start.\\

Human: English text: I have a headache.\\
Model-1: Başım ağrıyor.\\
Model-2: Başımda ağrı var.\texttt{>>}\\

\subsubsection{GPT-4-Turbo prompt}

\texttt{<<}System: You are a professional translator and interpreter specializing in healthcare and biomedical. You will compare and evaluate between two translations.\\

User: I want you to help me assess translations from English to Turkish. The translated texts are from the healthcare and biomedical domain. Your evaluation should focus on adherence to medical terminology, simplicity, clearness, and accuracy, and you are expected to assign a score ranging from 0 to 100 to each
translation. You MUST think, evaluate yourself, and do not generate text. Your ultimate output MUST only consist of the evaluation scores for each translation in JSON format. Do you understand your task?\\

Assistant: Certainly, I understood. You will provide two different translations from English to Turkish, and I will evaluate as a medical translator expert and assign scores to each one by adherence to medical terminology, simplicity, consistency and accuracy. The format will be a JSON file containing only scores. Ok, let us start.\\

User: English text: I have a headache.\\
Model-1: Baş ağrım var.\\
Model-2: Başımda ağrı var.\\
Assistant: \{"Model-1”: 90, "Model-2”: 80\}\texttt{>>}\\

The potential of Large Language Models (LLMs) in translation studies was evaluated as part of this research. A notable challenge when using LLMs for translation is their inability to provide fuzzy-match suggestions. To address this, we randomly selected 10,000 sentences from the medical domain and generated their embeddings using the "multilingual-e5-large" model \cite{huggingface2023}. These embeddings were stored using FAISS libraries. We then selected 725 sentences from our existing test set for prediction. During the prediction process, we retrieved the two sentences most similar to the input sentence from the pre-generated set of 10,000, based on cosine similarity scores, to create a fuzzy match. The performance scores of these LLM-based predictions are presented in Table \ref{tab:Table3}, \ref{tab:Table4}, \ref{tab:Table5}, and Table \ref{tab:Table6}.\\

For the development of our Machine Translation (MT) models, we prepared datasets in TMX, Moses, and Dublin Core formats. Data preparation involved several steps: removing HTML tags, decoding Unicode characters, and calculating character count ratios between sentence pairs. To mitigate anomalies, we selected sentence pairs with character count ratios between 0.8 and 1.25, considering the specific characteristics of each language. This process resulted in the creation of train, test, and development datasets, with the number of sentence pairs and tokens for each detailed in Table \ref{tab:Table2}.\\

Following data cleaning, sentence alignment, and corpus preparation based on predetermined ratios, we employed sentence transformers to assess the semantic relationships between sentence pairs. The semantic closeness of each pair was quantified using cosine similarity. We retained only those sentence pairs with a cosine similarity score of 0.90 or higher, thus finalizing our datasets through these filtration and transformation processes. Table \ref{tab:Table2} also presents the embeddings utilized in the development of our medical translation models.\\

The translation models were optimized using the following hyperparameters: a batch size of 16 for both training and evaluation, epoch sizes ranging from 20 to 50, and a learning rate of 2e-5. Model development involved training sessions lasting between 40 and 130 hours, depending on the specific language pair and dataset size.\\

\section{Result}
Extensive research in language translation reveals a continued demand for high-end translation services, particularly in the medical field. To meet this need, we have introduced a medical translation model covering six languages meticulously designed for use by healthcare professionals and various stakeholders.\\

This model performed remarkably well when evaluated against our in-house test datasets. The medical translation field currently needs more shared open-source benchmark test data. In conclusion, the findings presented by AI Amplified originate from our in-house test datasets, which have been precisely crafted in line with AI best practices and algorithmic principles.\\

The performance of our translation models has been rigorously evaluated using well-established benchmark scores such as BERT, BLEU, METEOR, and ROUGE Scores, which have been widely used in numerous studies (Tables \ref{tab:Table3}, \ref{tab:Table4}, \ref{tab:Table5}, and Table \ref{tab:Table6}).\\

\begin{table}[H]
    \centering
    \caption{Comparison of Results of Trained Language Models (BERT Scores)}
    %\textbf{Table 3: Comparison of Results of Trained Language Models (BERT Scores)} \\
    \renewcommand{\arraystretch}{1.2} % Satır yüksekliğini artır
    \setlength{\arrayrulewidth}{1.5pt} % Üst ve alt çizgi kalınlığını ayarla
    \setlength{\tabcolsep}{4pt} % Sütun genişliklerini daralt
    \fontsize{9}{11}\selectfont % Tablo yazı boyutunu 9 yapar
%    \small % Metin punto boyutunu 10 yapar
    \begin{tabular}{lrrrrr}
        \hline
        \textbf{Model/Languages} & \textbf{Aimped} & \textbf{DeepL} & \textbf{Google} & \textbf{Base-Helsinki} & \textbf{GPT4-Turbo} \\
        \hline
        de-to-en & 0.969 & 0.935 & 0.937 & 0.906 & 0.955 \\
        en-to-de & 0.928 & 0.920 & 0.916 & 0.884 & 0.886 \\
        en-to-es & 0.957 & 0.930 & 0.931 & 0.925 & 0.926 \\
        es-to-en & 0.981 & 0.935 & 0.939 & 0.930 & 0.967 \\
        en-to-fr & 0.950 & 0.940 & 0.948 & 0.932 & 0.937 \\
        fr-to-en & 0.976 & 0.946 & 0.950 & 0.934 & 0.969 \\
        en-to-pt & 0.954 & 0.945 & 0.947 & 0.940 & 0.938 \\
        pt-to-en & 0.979 & 0.947 & 0.951 & 0.972 & 0.972 \\
        en-to-ro & 0.964 & 0.921 & 0.924 & 0.931 & 0.851 \\
        ro-to-en & 0.993 & 0.953 & 0.955 & 0.950 & 0.973 \\
        en-to-tr & 0.886 & 0.892 & 0.882 & 0.826 & - \\
        tr-to-en & 0.959 & 0.928 & 0.926 & 0.876 & - \\
        \hline
    \end{tabular}
    \label{tab:Table3}
\end{table}

Analysis of the BERT and BLEU scores demonstrates that our models achieve highly satisfactory and statistically significant results. The strategy of fine-tuning with high-quality, domain-specific datasets has yielded exceptionally positive outcomes in specialized medical translation tasks. These results underscore the effectiveness of our approach in capturing the nuances of medical terminology and context.\\

\begin{table}[H]
    \centering
    \caption{Comparison of Results of Trained Language Models (BLEU Scores)}
    %\textbf{Table 4: Comparison of Results of Trained Language Models (BLEU Scores)} \\
    \renewcommand{\arraystretch}{1.2} % Satır yüksekliğini artır
    \setlength{\arrayrulewidth}{1.5pt} % Üst ve alt çizgi kalınlığını ayarla
    \setlength{\tabcolsep}{4pt} % Sütun genişliklerini daralt
    \fontsize{9}{11}\selectfont % Tablo yazı boyutunu 9 yapar
    %\small % Metin punto boyutunu 10 yapar
    \begin{tabular}{lrrrrr}
        \hline
        \textbf{Model/Languages} & \textbf{Aimped} & \textbf{DeepL} & \textbf{Google} & \textbf{Base-Helsinki} & \textbf{GPT4-Turbo} \\
        \hline
        de-to-en & 65.44 & 60.2 & 60.43 & 45.83 & 54.13 \\
        en-to-de & 56.49 & 52.59 & 49.77 & 36.93 & 44.3 \\
        en-to-es & 66.63 & 51.8 & 51.27 & 48.7 & 48.99 \\
        es-to-en & 69.6 & 55.36 & 56.19 & 52.17 & 52.12 \\
        en-to-fr & 62.15 & 56.8 & 60.52 & 51.62 & 56.49 \\
        fr-to-en & 66.3 & 62.02 & 63.07 & 52.54 & 58.4 \\
        en-to-pt & 62.03 & 55.73 & 56.98 & 51.85 & 51.52 \\
        pt-to-en & 66.89 & 58.16 & 61.05 & 58.2 & 54.59 \\
        en-to-ro & 75.04 & 48.1 & 50.19 & 54.49 & 38.06 \\
        ro-to-en & 89.71 & 61.5 & 62.58 & 59.37 & 58.06 \\
        en-to-tr & 42.99 & 45.06 & 42.38 & 28.83 & - \\
        tr-to-en & 48.86 & 50.62 & 49.37 & 29.2 & - \\
        \hline
    \end{tabular}    
    \label{tab:Table4}
\end{table}

Despite the relatively modest size of our datasets, our models achieve notable performance due to the high quality of the curated sentence pairs. This underscores the importance of data quality over quantity in developing effective translation models. Moreover, the efficiency of our approach is highlighted by the short fine-tuning times required, which not only accelerates model development but also suggests potential for rapid adaptation to new medical subdomains or emerging terminology\\

\begin{table}[H]
    \centering
    \caption{Comparison of Results of Trained Language Models (METEOR)}
    % \textbf{Table 5: Comparison of Results of Trained Language Models (METEOR)} \\
    \renewcommand{\arraystretch}{1.2} % Satır yüksekliğini artır
    \setlength{\arrayrulewidth}{1.5pt} % Üst ve alt çizgi kalınlığını ayarla
    \setlength{\tabcolsep}{4pt} % Sütun genişliklerini daralt
    \fontsize{9}{11}\selectfont % Tablo yazı boyutunu 9 yapar
    %\small % Metin punto boyutunu 9 yapar
    \begin{tabular}{lrrrrr}
        \hline
        \textbf{Model/Languages} & \textbf{Aimped} & \textbf{DeepL} & \textbf{Google} & \textbf{Base-Helsinki} & \textbf{GPT4-Turbo} \\
        \hline
        de-to-en & 0.835 & 0.812 & 0.815 & 0.716 & 0.794 \\
        en-to-de & 0.792 & 0.758 & 0.748 & 0.677 & 0.690 \\
        en-to-es & 0.853 & 0.756 & 0.757 & 0.739 & 0.739 \\
        es-to-en & 0.882 & 0.800 & 0.811 & 0.782 & 0.793 \\
        en-to-fr & 0.827 & 0.785 & 0.816 & 0.764 & 0.790 \\
        fr-to-en & 0.867 & 0.824 & 0.853 & 0.800 & 0.831 \\
        en-to-pt & 0.829 & 0.792 & 0.797 & 0.769 & 0.771 \\
        pt-to-en & 0.872 & 0.831 & 0.849 & 0.833 & 0.829 \\
        en-to-ro & 0.869 & 0.722 & 0.732 & 0.759 & 0.575 \\
        ro-to-en & 0.956 & 0.834 & 0.848 & 0.830 & 0.827 \\
        en-to-tr & 0.686 & 0.699 & 0.675 & 0.553 & - \\
        tr-to-en & 0.769 & 0.775 & 0.779 & 0.600 & - \\
        \hline
    \end{tabular}
    \label{tab:Table5}
\end{table}

Our models demonstrate consistently high performance across multiple evaluation metrics. In addition to the strong BERT and BLEU scores previously mentioned, the METEOR and ROUGE scores also indicate exceptional translation quality. These results collectively underscore the robustness of our approach. However, it is crucial to note a significant limitation in our evaluation process. Due to the scarcity of standardized, open-source test sets specifically designed for benchmarking medical translation models, all reported scores are derived from test sets based on our proprietary in-house datasets. While this allows for a consistent evaluation across our models, it may limit direct comparability with other studies or systems using different datasets.\\

This limitation underscores the need for the development and adoption of standardized, publicly available test sets in the field of medical translation. Such resources would greatly enhance the ability to compare different approaches and models across the research community\footnote[2]{ \url{https://github.com/ai-amplified/models/tree/main/medical_translation/test_data}}.

\begin{table}[H]
    \centering
    \caption{Comparison of Results of Trained Language Models (ROUGE Scores)}
    %\textbf{Table 6: Comparison of Results of Trained Language Models (ROUGE Scores)} \\
    \renewcommand{\arraystretch}{1.1} % Satır yüksekliğini artır
    \setlength{\arrayrulewidth}{1.5pt} % Üst ve alt çizgi kalınlığını ayarla
    \setlength{\tabcolsep}{4pt} % Sütun genişliklerini daralt
    \fontsize{9}{11}\selectfont % Tablo yazı boyutunu 9 yapar
    %\small % Metin punto boyutunu 9 yapar
    \begin{tabular}{lrrrrr}
        \hline
        \textbf{Model/Languages} & \textbf{Aimped} & \textbf{DeepL} & \textbf{Google} & \textbf{Base-Helsinki} & \textbf{GPT4-Turbo} \\
        \hline
        de-to-en & 0.841 & 0.817 & 0.822 & 0.723 & 0.769 \\
        en-to-de & 0.798 & 0.765 & 0.758 & 0.673 & 0.668 \\
        en-to-es & 0.866 & 0.775 & 0.776 & 0.763 & 0.732 \\
        es-to-en & 0.886 & 0.812 & 0.819 & 0.790 & 0.768 \\
        en-to-fr & 0.847 & 0.809 & 0.842 & 0.796 & 0.790 \\
        fr-to-en & 0.871 & 0.830 & 0.858 & 0.807 & 0.810 \\
        en-to-pt & 0.850 & 0.814 & 0.825 & 0.779 & 0.774 \\
        pt-to-en & 0.877 & 0.834 & 0.855 & 0.809 & 0.804 \\
        en-to-ro & 0.897 & 0.804 & 0.809 & 0.810 & 0.773 \\
        ro-to-en & 0.961 & 0.849 & 0.855 & 0.841 & 0.803 \\
        en-to-tr & 0.742 & 0.752 & 0.733 & 0.626 & - \\
        tr-to-en & 0.787 & 0.790 & 0.790 & 0.634 & - \\
        \hline
    \end{tabular}
    \label{tab:Table6}
\end{table}

AI-Amplified's translation models demonstrate exceptional performance across multiple evaluation metrics. The scores obtained for BERT, BLEU, METEOR, and ROUGE consistently rank among the highest in our comparative analysis, with only a few minor exceptions. This comprehensive evaluation underscores the robustness and accuracy of our models in medical translation tasks.\\

\begin{figure}[H]
    \centering
    \fbox{\includegraphics[width=0.7\linewidth]{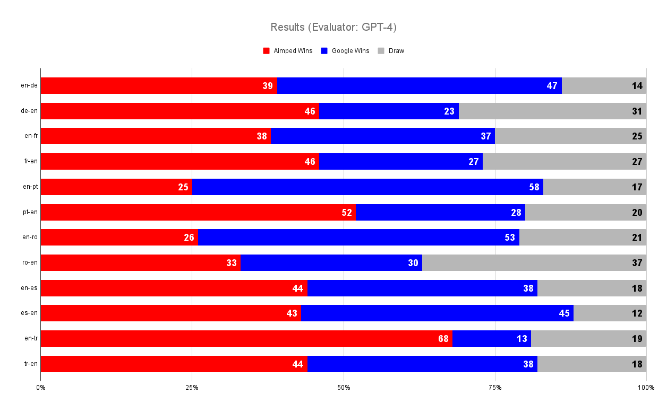}}
    \caption{Refereed by GPT-4-Turbo Results}
    \label{fig:figure_page8_1}
\end{figure}

\begin{figure}[H]
    \centering
    \fbox{\includegraphics[width=0.7\linewidth]{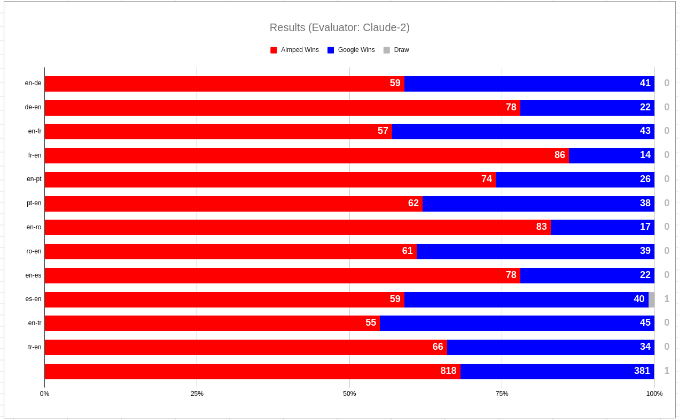}}
    \caption{Refereed by Claude-3-Opus Results}
    \label{fig:figure_page8_2}
\end{figure}

In this work, we also compared the results of Google Translate and our MT model [53] in six languages using two LLM models, GPT4-Turbo (Figure \ref{fig:figure_page8_1} and Table\ref{tab:Table7}) and Claude-3-Opus (Figure \ref{fig:figure_page8_2} and Table\ref{tab:Table8}), as referees.\\

In the MT comparison for GPT4-Turbo with a total of 1200 instances, AI-Amplified outperformed Google Translate in 504 instances, while Google Translate performed better in 437 instances, and both models achieved the same score in 259 instances.\\

\begin{table}[h]
    \centering
    \caption{AI-Amplified MT \& Google Translate Translation Comparison Results (Refereed by GPT4-Turbo)}
    %\textbf{Table 7: AI Amplified MT \& Google Translate Translation Comparison Results (Refereed by GPT4-Turbo)} \\
    \renewcommand{\arraystretch}{1.1} % Satır yüksekliğini artır
    \setlength{\arrayrulewidth}{1.5pt} % Üst ve alt çizgi kalınlığını ayarla
    \setlength{\tabcolsep}{4pt} % Sütun genişliklerini daralt
    \fontsize{9}{11}\selectfont % Tablo yazı boyutunu 9 yapar
    %\small % Metin punto boyutunu 9 yapar
    \begin{tabular}{lccc}
        \hline
        \textbf{MT Models \& Languages} & \textbf{Aimped} & \textbf{Google Translate} & \textbf{Draw} \\
        \hline
        en-de & 39 & 47 & 14 \\
        de-en & 46 & 23 & 31 \\
        en-fr & 38 & 37 & 25 \\
        fr-en & 46 & 27 & 27 \\
        en-pt & 25 & 58 & 17 \\
        pt-en & 52 & 28 & 20 \\
        en-ro & 26 & 53 & 21 \\
        ro-en & 33 & 30 & 37 \\
        en-es & 44 & 38 & 18 \\
        es-en & 43 & 45 & 12 \\
        en-tr & 68 & 13 & 19 \\
        tr-en & 44 & 38 & 18 \\
        \hline
        \textbf{Total} & \textbf{504} & \textbf{437} & \textbf{259} \\
        \hline
    \end{tabular}
    \label{tab:Table7}
\end{table}

In the context of language samples, using GPT4-Turbo as a referee, our models produce better scores than Google Translate models in most translations.\\

In the MT comparison for Claude-3-Opus with a total of 1200 samples, AI-Amplified outperformed Google Translate in 818 samples, while Google Translate outperformed in 381 samples, with both models achieving the same score in only one sample.

\begin{table}[h]
    \centering
    \caption{AI-Amplified MT \& Google Translate Translation Comparison Results (Refereed by Claude-3-Opus)}
    % \textbf{Table 8: AI Amplified MT \& Google Translate Translation Comparison Results (Refereed by Claude-3-Opus)} \\
    %\renewcommand{\arraystretch}{1.5} % Satır yüksekliğini artır
    \setlength{\arrayrulewidth}{1.5pt} % Üst ve alt çizgi kalınlığını ayarla
    \setlength{\tabcolsep}{4pt} % Sütun genişliklerini daralt
    \fontsize{9}{11}\selectfont % Tablo yazı boyutunu 9 yapar
%    \small % Metin punto boyutunu 9 yapar
    \begin{tabular}{lccc}
        \hline
        \textbf{MT Models \& Languages} & \textbf{Aimped} & \textbf{Google Translate} & \textbf{Draw} \\
        \hline
        en-de & 59 & 41 & 0 \\
        de-en & 78 & 22 & 0 \\
        en-fr & 57 & 43 & 0 \\
        fr-en & 86 & 14 & 0 \\
        en-pt & 74 & 26 & 0 \\
        pt-en & 62 & 38 & 0 \\
        en-ro & 83 & 17 & 0 \\
        ro-en & 61 & 39 & 0 \\
        en-es & 78 & 22 & 0 \\
        es-en & 59 & 40 & 1 \\
        en-tr & 55 & 45 & 0 \\
        tr-en & 66 & 34 & 0 \\
        \hline
        \textbf{Total} & \textbf{818} & \textbf{381} & \textbf{1} \\
        \hline
    \end{tabular}
    \label{tab:Table8}
\end{table}

In the context of language samples, using Claude-3-Opus, as a referee, our models produce better scores than Google Translate models in all translations. Sentence pairs of our MT models developed by AI-Amplified are available on our GitHub page \footnote[3]{ \url{https://github.com/ai-amplified}}.\\

To investigate transfer learning methods of clinical texts with multilingual pre-trained language models (MPLMs) in a study evaluating the performance of medical text translations between English and Spanish \cite{han2023}, Clinical-Marian \cite{junczys-dowmunt2018}, Clnical-NLLB \cite{costa-jussa2022} and Clinical-WMT21fb \cite{tran2021} models were compared. The ClinSpEn-2022\footnote[4]{ \url{https://temu.bsc.es/clinspen/}} competition dataset was used for the evaluation. BLUE, METEOR, ROUGE, and COMET scores were compared. The ClinSpEn-2022 dataset used in this study was evaluated as a benchmark and the scores we obtained as AI Amplified were compared with those obtained by Han, Gladkoff \cite{han2023} (Table \Ref{tab:Table9}).\\

\begin{table}[H] % H seçeneği tablonun tam olarak burada olmasını sağlar
    \centering
    \caption{Comparison of Translation Model Performance (ClinEspen 22 benchmark)}
    %\textbf{Table 9: Comparison of Translation Model Performance (ClinEspen 22 benchmark)} \\
    \renewcommand{\arraystretch}{1.2} % Satır yüksekliğini artır
    \setlength{\arrayrulewidth}{1.2pt} % Üst ve alt çizgi kalınlığını ayarla
    \setlength{\tabcolsep}{4pt} % Sütun genişliklerini daralt
    \fontsize{9}{11}\selectfont % Metin punto boyutunu 9 yapar, satır aralığını 11 yapar
    \begin{tabular}{lcccccc}
        %\toprule % Kalın üst çizgi
        \multicolumn{1}{c}{\fontsize{10}{12}\selectfont \textbf{}} & \multicolumn{1}{c}{\fontsize{10}{12}\selectfont \textbf{SACREBLEU}} & \multicolumn{1}{c}{\fontsize{10}{12}\selectfont \textbf{METEOR}} & \multicolumn{1}{c}{\fontsize{10}{12}\selectfont \textbf{COMET}} & \multicolumn{1}{c}{\fontsize{10}{12}\selectfont \textbf{BLEU}} & \multicolumn{1}{c}{\fontsize{10}{12}\selectfont \textbf{ROUGE-L-F1}} \\
        \hline
        \multicolumn{6}{c}{\textbf{Task-I: Clinical Cases (CC) EN→ES}} \\
        MT-fine-tuning Models & & & & & \\
        Clinical-Marian & 38.18 & 0.6338 & 0.4237 & 0.365 & 0.6271 \\
        Clinical-NLLB & 37.74 & 0.6273 & 0.4081 & 0.3601 & 0.6193 \\
        Clinical-WMT21fb & 34.3 & 0.5868 & 0.3448 & 0.3266 & 0.5927 \\
        Aimped & 38.43 & 0.6444 & - & 0.366 & 0.6288 \\
        \hline
        \multicolumn{6}{c}{\textbf{Task-II: Clinical Terms (CT) EN←ES}} \\
        MT-fine-tuning & & & & & \\
        Clinical-Marian & 26.87 & 0.5885 & 0.9791 & 0.2667 & 0.672 \\
        Clinical-NLLB & 28.57 & 0.5873 & 1.029 & 0.2844 & 0.671 \\
        Clinical-WMT21fb & 24.39 & 0.584 & 0.8584 & 0.2431 & 0.6699 \\
        Aimped & 37.85 & 0.6141 & - & 0.3777 & 0.7146 \\
        \hline
        \multicolumn{6}{c}{\textbf{Task-III: Ontology Concept (OC) EN→ES}} \\
        MT-fine-tuning & & & & & \\
        Clinical-Marian & 39.1 & 0.6262 & 0.9495 & 0.3675 & 0.7688 \\
        Clinical-NLLB & 41.63 & 0.6072 & 0.918 & 0.3932 & 0.7477 \\
        Clinical-WMT21fb & 40.71 & 0.5686 & 0.9908 & 0.3859 & 0.7199 \\
        Aimped & 44.14 & 0.579 & - & 0.4302 & 0.7283 \\
      %  \bottomrule % Kalın alt çizgi
      \hline
    \end{tabular}
    \label{tab:Table9}    
\end{table}

The results show that the MT model we developed as AI-Amplified scores above the scores obtained by Han, Gladkoff \cite{han2023} with the relevant benchmark dataset in most cases. Han, Gladkoff \cite{han2023} state that the Marian model is much smaller than the other two models. However, the Marian model surpassed these two models in terms of the score it obtained. As a result, it is revealed that models with many more parameters do not always produce good scores. Data quality and fine-tuning may be more important than model size alone. Therefore, it is shown that LLMs may not necessarily be better and that the quality of the data set and training is also essential.\\

\section{Conclusion}
Machine Translation (MT) has become increasingly crucial in artificial intelligence due to globalization and technological advancements. The need for accurate translations, particularly in sharing scientific research and medical terminology across languages, has grown exponentially. This study demonstrates that AI-based MT, specifically tailored for the medical domain, plays a critical role in meeting this demand. Our research introduced a novel "LLMs-in-the-loop" approach to develop specialized neural machine translation models for medical texts. We focused on six language pairs: English to and from Spanish, German,
French, Romanian, Turkish, and Portuguese. The models employ encoder-decoder architectures with LSTM units, trained on custom parallel corpora compiled from scientific articles and medical texts.\\

Key findings of our study include:\\

1. Small, specialized models trained on high-quality in-domain data can outperform larger, generalpurpose models. For instance, our English-German model achieved BERT and BLEU Scores of 0.969 and 65.44, surpassing GPT-4 Turbo's scores of 0.955 and 54.13 respectively.\\

2. The LLMs-in-the-loop methodology, incorporating synthetic data generation, rigorous model evaluation, and agent orchestration, significantly enhanced model performance.\\

3. When evaluated by LLM models like GPT-4 and Claude-3 acting as expert judges, our models were preferred over Google Translate in 68-86\% of sample evaluations across all languages.\\

4. Balancing domain-specific data with out-of-domain data (10-20\% ratio) during fine-tuning helps maintain translation quality for general expressions while excelling in medical terminology.\\

5. The importance of starting with a strong general translation model before fine-tuning with domainspecific data was highlighted.\\

Our approach of combining medical domain-specific datasets with a proportion of out-of-domain data yielded successful results. However, we noted that fine-tuning a general language model with purely domain-specific data can decrease translation quality for out-of-domain texts. Our method of merging domain-specific and
general datasets mitigated this issue.\\

While our results are promising, it's important to note that they were obtained using our proprietary datasets due to limited availability of open-source medical translation test sets. This underscores the need for more publicly available resources in this field.\\

Looking ahead, the development of AI models working with more languages and larger datasets holds great potential for advancing medical translation. As AI-Amplified, we are committed to continually improving our datasets and models to achieve higher performance scores.\\

This study not only demonstrates the effectiveness of our specialized MT models in the medical field but also lays the groundwork for future developments in healthcare-related natural language processing, including planned models for deidentification and healthcare entity extraction.\\

Our medical translation models are accessible on our website \cite{ai-amplified2023}, where demo translations can be tested. We believe these tools will significantly contribute to the global health community by facilitating knowledge sharing and accelerating scientific communication across language barriers.

\end{document}